%% file: camera.tex
\documentclass[runningheads]{llncs}

\usepackage{eccv}

\usepackage{eccvabbrv}

\usepackage{graphicx}
\usepackage{booktabs}
\usepackage{multirow}
\usepackage[table]{xcolor}
\usepackage[pagebackref,breaklinks,colorlinks,citecolor=eccvblue]{hyperref}

\usepackage[accsupp]{axessibility}  

\usepackage{hyperref}

\usepackage{orcidlink}

\begin{document}

\title{RGB-Pointmap Pretraining for Unified 3D Scene Understanding}
\titlerunning{RGB-Pointmap Pretraining}

\author{Ye Mao, Weixun Luo, Ranran Huang, Junpeng Jing\thanks{Corresponding author: \email{j.jing23@imperial.ac.uk}}, Krystian Mikolajczyk }

\institute{Imperial College London \\
\url{https://yebulabula.github.io/UniScene3D/} 
}

\authorrunning{Y. Mao et al.}



\maketitle
\input{figures/teaser}
\begin{abstract}
Pretraining 3D encoders through alignment with Contrastive Language–Image Pre-training (CLIP) has emerged as a promising direction to learn generalizable representations for 3D scene understanding. In this paper, we propose UniScene3D, a transformer-based framework that learns unified 3D scene representations from multi-view RGB–Pointmap by leveraging the priors of a pretrained 2D foundation model. For robust RGB-Pointmap representation learning, we introduce novel cross-view geometric alignment and grounded view alignment to enforce geometry and semantic consistency across views. Extensive low-shot and task-specific fine-tuning across viewpoint grounding, scene retrieval, scene classification, and 3D visual question answering achieves state-of-the-art performance. These results establish UniScene3D as an effective framework for unified 3D scene understanding.

\end{abstract}

\section{Introduction}
\label{sec:intro}
3D scene understanding, covering tasks such as 3D segmentation~\cite{he2021deep,tchapmi2017segcloud} and 3D visual question answering (3D VQA)~\cite{azuma2022scanqa,wang20233d}, is a fundamental capability for embodied agents and VR/AR systems~\cite{li2025embodied,duan2022survey}. Learning generalizable 3D representations that can be transferred across these scene tasks has been a long-standing research topic. Recently, an emerging 3D representation learning paradigm is to align 3D encoders with pretrained vision–language models such as CLIP~\cite{radford2021learning,tschannen2025siglip}.

Within this paradigm, recent works explore different input modalities for training CLIP-aligned 3D encoders, including point clouds~\cite{zhouuni3d,liu2023openshape,xue2023ulip,fan2024point}, multi-view images~\cite{dong2024simc3d,lee2024duoduo,zhou2025cross}, depth maps~\cite{mao2024opendlign,chen2023clip2scene}, and pointmaps~\cite{mao2025poma}. Each modality offers distinct advantages but also exhibits inherent limitations. Point clouds capture explicit 3D geometry, but their irregular and unordered structure makes them incompatible with grid-based architectures, limiting the effective use of pretrained 2D priors. Multi-view images and depth maps preserve a regular grid structure, yet they lack globally consistent 3D geometry at the scene level. This is because each view is defined in its individual image or camera frame. In contrast, pointmaps utilize a shared world frame to encode coherent 3D coordinates while preserving an image-like grid representation, making them a promising modality for 3D representation learning with standard 2D backbones.

Despite its potential, research on pointmap-based representation learning remains in an early stage, and several key challenges remain underexplored. First, pointmap itself only has geometric coordinates and lacks appearance cues such as color and texture. It limits their ability to capture visually grounded semantics, especially in 3D understanding scenarios that require reasoning over visual attributes (see Fig.~\ref{fig:figure1}). Second, the geometric consistency across multiple pointmap views provides a natural signal for cross-view representation learning. This encourages robust scene representations that overcome the partial observability and view-dependent occlusions inherent to single-view modeling. Conventional multi-view learning strategies~\cite{dong2024simc3d,lee2024duoduo,zhou2025cross,mao2024opendlign} process views independently and thus struggle to fully exploit this structural property.


To address these challenges, we introduce UniScene3D, a 3D transformer encoder that learns \textbf{Uni}fied \textbf{Scene} representations jointly capturing appearance and geometry from multi-view image–pointmap pairs, which we refer to as \textit{\textbf{RGB-Pointmap}}. Specifically, we propose an early fusion module that integrates image and pointmap tokens at the patch embedding stage, enabling a single encoder to process both modalities. We further design multimodal alignment objectives to explicitly capture the cross-view relations of RGB-Pointmap. In particular, cross-view geometric alignment enforces similarity between spatially proximate views and distinction for distant ones. Grounded view alignment associates an object referring text with the views that observe the object, thereby encouraging aligned representation for views with shared object semantics.

The effectiveness of our method is validated across multiple 3D scene understanding tasks, including viewpoint grounding, scene retrieval, scene type classification, and 3D VQA. These tasks probe representation quality from object-level and view-level reasoning to holistic scene understanding, with potential applications in robotic navigation, embodied perception, and human–robot interaction~\cite{chen2024scene,sarkar2025crossover}. Experimental results show that UniScene3D consistently outperforms state-of-the-art vision encoders across all tasks (see Fig.~\ref{fig:figure1}) under zero-shot, few-shot, and task-specific fine-tuning settings. These results highlight that our model effectively encodes both geometric and appearance cues in RGB-Pointmap to produce generalizable 3D representations. The key contributions are summarized below:
\begin{itemize}
    \item We propose UniScene3D, a vision encoder that learns unified 3D scene representations by jointly modeling geometry and appearance from multi-view RGB-Pointmap.
    \item We introduce novel alignment objectives, including cross-view geometric alignment and grounded view alignment, to capture geometric and semantic consistency in RGB-Pointmap across viewpoints.
    \item We evaluate our method across diverse downstream 3D scene understanding tasks under zero-shot, few-shot, and task-specific fine-tuning settings, achieving state-of-the-art performance.
\end{itemize}

\section{Related Work}
\label{sec:relatedwork}
In this section, we first review existing 3D representation learning methods based on vision–language pretraining, and then summarize commonly used 3D scene datasets for pretraining and the evaluation protocols for vision–language models.

\noindent \textbf{3D Vision–Language Pretraining. } 
3D vision–language pretraining aligns a 3D encoder with pretrained CLIP models~\cite{radford2021learning,tschannen2025siglip,sun2023eva,fang2023data} and has become a common paradigm for 3D representation learning. Most previous works adopt point clouds as the input modality to train encoders. PointCLIP-series~\cite{zhang2022pointclip,zhu2023pointclip}, ULIP-series~\cite{xue2023ulip,xue2024ulip}, PointBind~\cite{guo2023point}, 
CLIP$^2$~\cite{zeng2023clip2}
OpenShape~\cite{liu2023openshape}, Uni3D~\cite{zhouuni3d}, MixCon3D~\cite{gao2023mixcon3d}, and OccTIP~\cite{nguyen2025occlusion} pretrain point cloud encoders that show strong zero-shot performance on 3D object recognition and cross-modal 3D object retrieval. Subsequent work extends this paradigm to scene-level understanding~\cite{zhu20233d,jia2024sceneverse,arnaud2025locate,chen2023clip2scene,ding2023pla,yang2024regionplc}. However, point cloud encoders operate in a modality misaligned with 2D pretrained backbones, preventing direct reuse of rich 2D priors and requiring large-scale 3D pretraining to maintain generalization.

To better exploit 2D pretrained priors, alternative approaches adopt 2D-compatible representations. DuoduoCLIP~\cite{lee2024duoduo} uses multi-view images, while OpenDlign~\cite{mao2024opendlign} and CLIP2Point~\cite{huang2023clip2point} use depth maps to adapt 2D encoders to 3D tasks. However, these projections retain incomplete geometric information, which limits performance on complex and global spatial reasoning. Recent methods based on 3D Gaussian splatting~\cite{qin2024langsplat,li2025langsplatv2,jiao2025clip,liao2025clip,li2025scenesplat} provide another direction, but are primarily evaluated on 3D segmentation, restricting their demonstrated applicability. POMA-3D~\cite{mao2025poma} is an early attempt at learning 3D representations from pointmaps. However, POMA-3D relies solely on pointmaps and thus misses important appearance cues. Its multi-stage training and JEPA-style cross-view alignment also incur substantial training costs. In contrast, our method jointly models geometry and appearance from images and pointmaps to learn unified 3D representations. The proposed geometric and grounded view alignment objectives further enable UniScene3D to exploit cross-view information more effectively while remaining computationally efficient. 

\noindent \textbf{3D Scene Pretraining Datasets.}
Collecting accurate 3D scene data at scale remains challenging due to the high cost and time-intensive nature of 3D scanning and reconstruction~\cite {hou2021exploring,yao2020blendedmvs,bi2010advances,feng2024shape2scene,huang2026none,schonberger2016structure,schonberger2016structure}. Existing real-world datasets, including ScanNet~\cite{dai2017scannet}, 3RScan~\cite{wald2019rio}, ARKitScenes~\cite{baruch2021arkitscenes}, HM3D~\cite{ramakrishnan2021habitat}, and MultiScan~\cite{mao2022multiscan}, contain only thousands of scenes, orders of magnitude fewer than the billions of images used for 2D pretraining. Some works~\cite{jia2024sceneverse} mitigate this limitation by incorporating synthetic datasets~\cite{zheng2020structured3d, deitke2022️}. However, the domain gap between synthetic and real-world environments often constrains transferability. Accordingly, many 3D pretraining methods~\cite{zhouuni3d,liu2023openshape,xue2023ulip} transfer knowledge from large-scale pretrained vision models using limited 3D data. Following this principle, UniScene3D is post-trained from a pretrained 2D image encoder and leverages thousands of real-world 3D scenes to learn transferable 3D representations.

\noindent \textbf{Evaluation Protocol for Vision–Language Models.}
Low-shot evaluation (i.e., zero-shot and few-shot) provides a standardized protocol for assessing vision–language pretrained encoders by measuring representation quality under minimal supervision, thereby isolating encoder capacity. In the 2D domain, CLIP models~\cite{radford2021learning,fang2023data,zhai2023sigmoid,tschannen2025siglip,sun2023eva} are commonly evaluated through low-shot image classification. In 3D, object-centric models~\cite{xue2023ulip,liu2023openshape,zhouuni3d,mao2024opendlign,zhu2023pointclip} are typically assessed via low-shot 3D object classification and retrieval.

In contrast, most 3D scene representation models are evaluated only under task-specific fine-tuning on benchmarks~\cite{azuma2022scanqa,ma2022sqa3d,mao2025hypo3d,chen2020scanrefer,chen2021scan2cap,dai2017scannet,yeshwanth2023scannet++} that focus on high-level reasoning or dense prediction tasks, such as 3D VQA, grounding, detection, captioning, and segmentation. To address the lack of systematic low-shot evaluation, we additionally assess our model on several low-shot tasks, including viewpoint grounding, scene retrieval, and scene classification, enabling a more comprehensive assessment of the scene representation quality. \input{figures/uniscene3d}

\section{UniScene3D}
\label{sec:uniscene3d}
UniScene3D is a 3D vision–language pretraining framework that learns unified 3D scene representations from multi-view image–pointmap pairs (Fig.~\ref{fig:2}). In Sec.~\ref{sec3.1}, we describe the model architecture for constructing RGB-Pointmap representations. In Sec.~\ref{sec3.2}, we introduce four multimodal alignment objectives that jointly enable robust 3D feature learning.

\subsection{RGB-Pointmap Representation}\label{sec3.1}
To effectively integrate appearance and geometric information, the input to UniScene3D consists of RGB-Pointmap represented as $V$ aligned multi-view image–pointmap pairs ${(\mathbf{I}_v, \mathbf{P}_v)}_{v=1}^{V}$. For each view $v$, $\mathbf{I}_v \in \mathbb{R}^{H \times W \times 3}$ denotes the RGB image, and $\mathbf{P}_v \in \mathbb{R}^{H \times W \times 3}$ denotes the pixel-aligned pointmap expressed in the world coordinate frame. Each pointmap $\mathbf{P}_v$ is constructed by back-projecting the corresponding depth map using the camera intrinsics and extrinsics.

Built upon and initialized from the pretrained 2D image encoder of FG-CLIP~\cite{xie2025fg}, we introduce an early fusion strategy at the patch embedding stage to encode RGB-Pointmap. After patchification of each view, the image patch embedding layer $\phi_I$ projects $M$ image patches of $\mathbf{I}_v$ into $\mathbf{z}^I_v \in \mathbb{R}^{M \times d}$, formulated as $\mathbf{z}^I_v = \phi_I(\mathbf{I}_v) + \mathbf{E}_{pos}$, where $\mathbf{E}_{pos} \in \mathbb{R}^{M \times d}$ denotes learnable positional encodings. In parallel, a pointmap patch embedding layer $\phi_P$, sharing the same architecture and initialization as $\phi_I$, maps geometric patches into $\mathbf{z}^P_v = \phi_P(\mathbf{P}_v) \in \mathbb{R}^{M \times d}$.
The two modalities of patch tokens are then fused via element-wise summation, $\mathbf{z}_v = \mathbf{z}^I_v + \mathbf{z}^P_v$. In this manner, appearance and geometric cues are integrated at the earliest representation stage, enabling joint reasoning across modalities while preserving architectural compatibility with the 2D backbone. We compare this approach with other fusion choices in the experimental section.

Then, a learnable class token $\mathbf{z}^{cls} \in \mathbb{R}^d$ is replicated across views as $\mathbf{z}_v^{cls} = \mathbf{z}^{cls}$. The concatenation of the per-view class token and the fused patch tokens $[\mathbf{z}_v^{cls}; \mathbf{z}_v] \in \mathbb{R}^{(M+1)\times d}$ is subsequently processed by $N$ transformer blocks.
We take the output per-view class token as per-view representation, denoted as $\mathbf{h}_v$ at view $v$, then the RGB-Pointmap representation of the entire scene is therefore defined as the set of per-view representation $\{\mathbf{h}_v\}_{v=1}^{V}$.

\input{tables/comparison_of_alignment}

\subsection{Multimodal Alignment}\label{sec3.2}
To pretrain UniScene3D, we employ four multimodal alignment objectives. As summarized in Table~\ref{tab:2}, cross-view geometric alignment and grounded view alignment construct contrastive pairs within each scene to enforce cross-view consistency among pointmap view embeddings. View-level and scene-level alignments form view and scene pairs across the training batch, encouraging the model to learn discriminative representations across different scenes.

\noindent\textbf{Cross-view Geometric Alignment.}
Learning effective RGB-Pointmap representations requires accurately encoding the underlying 3D geometry of a scene across viewpoints. An embedding space organized by geometric overlap should place adjacent views closer than distant ones, enabling the representation to reflect spatial relationships within the scene. To this end, we introduce a cross-view geometric alignment objective that enforces embedding similarities to correspond to the spatial proximity of views within the same scene.

Consider two views $v$ and $u$ within a scene, whose pointmaps $\mathbf{P}_v=\{\mathbf{x}_i\}_{i=1}^{N}$ and $\mathbf{P}_u=\{\mathbf{y}_i\}_{i=1}^{N}$ each consist of $N$ 3D points, where $N=H\times W$, $\mathbf{x} \in \mathbb{R}^3$ and $\mathbf{y} \in \mathbb{R}^3$. The cross-view geometric dissimilarity is measured using the symmetric Chamfer distance~\cite{fan2017point}:
\begin{equation}
\mathrm{CD}(\mathbf{P}_v,\mathbf{P}_u)
=
\frac{1}{N}\sum_{\mathbf{x}\in P_v}
\min_{\mathbf{y}\in P_u}\|\mathbf{x}-\mathbf{y}\|_2^2
+
\frac{1}{N}\sum_{\mathbf{y}\in P_u}
\min_{\mathbf{x}\in P_v}\|\mathbf{y}-\mathbf{x}\|_2^2.
\end{equation}
For each anchor view $v$, all other views in the scene are sorted by increasing $\mathrm{CD}(\mathbf{P}_v,\mathbf{P}_u)$ to obtain an ordinal proximity index $r_v(u) \in \{0,1,\dots\}$,
where smaller values indicate stronger geometric proximity. For each anchor view $v$, the contrastive candidate set consists of all remaining views in the same scene, $\mathcal{V}_S \setminus {v}$, where $\mathcal{V}_S$ denotes the set of all views within that scene. We construct a rank-aware soft target distribution over this set:
\begin{equation}
p^{\text{soft}}_{v,u}
=
\frac{\exp\!\left(-r_v(u)/\tau_r\right)}
{\sum_{k \in \mathcal{V}_S \setminus \{v\}}
\exp\!\left(-r_v(k)/\tau_r\right)},
\end{equation}
where $\tau_r>0$ controls the concentration of probability mass over geometrically closer views. To stabilize optimization, this distribution is interpolated with a hard nearest-neighbor target $p^{\text{hard}}_{v,u}$:
\begin{equation}
p_{v,u}
=
\alpha\,p^{\text{hard}}_{v,u}
+
(1-\alpha)\,p^{\text{soft}}_{v,u},
\end{equation}
where $p^{\text{hard}}_{v,u}=1$ if $u=\arg\min_{k\neq v}\mathrm{CD}(\mathbf{P}_v,\mathbf{P}_k)$ and $0$ otherwise. The interpolation coefficient $\alpha\in[0,1]$ controls the contribution of the hard and soft targets: when $\alpha=1$, the objective reduces to hard nearest-neighbor contrastive learning, whereas $\alpha=0$ yields a fully rank-aware geometric alignment formulation. 
We define cross-view similarity logits as
$\ell_{v,u} = \mathbf{h}_v^\top \mathbf{h}_u/\tau$,
where $\tau>0$ is a learnable temperature parameter.
The geometric alignment loss is then defined as the soft-label cross-entropy over the candidate set:
\begin{equation}
\mathcal{L}_{\text{geo}}
=
-
\sum_{v}
\sum_{u \in \mathcal{V}_S \setminus \{v\}}
p_{v,u}
\log
\frac{\exp(\ell_{v,u})}
{\sum_{k \in \mathcal{V}_S \setminus \{v\}}
\exp(\ell_{v,k})}.
\end{equation}

\noindent\textbf{Grounded View Alignment.}
While cross-view geometric alignment enforces spatial consistency across viewpoints, RGB-Pointmap also exhibit cross-view semantic relationships because different views may observe the same objects. To capture this property, we introduce grounded view alignment, which aligns object referring text embeddings with all per-view RGB-Pointmap embeddings that observe the referred objects. This objective encourages views sharing the same object semantics to be closer in the representation space.

Let $\mathbf{t}_o \in \mathbb{R}^d$ denote the embedding of the referring text associated with a grounded object $o$. Positive view–object pairs are determined based on geometric visibility. Specifically, for each object $o$, we examine whether there is an intersection between each view 3D pointmap $\mathbf{P}_v$ and the object's 3D mask from the pretrained SceneVerse~\cite{jia2024sceneverse} dataset. If the intersection exists, view $v$ is considered to observe object $o$, and the pair $(v,o)$ is included in the set of valid positives $\mathcal{P}$. A single scene may thus produce multiple valid view–object correspondences.

For each $(v,o)\in\mathcal{P}$, we compute cross-modal similarity logits $s_{v,o}=\mathbf{h}_v^\top \mathbf{t}_o/{\tau}$. The grounded view alignment loss is computed over all valid positive pairs:
\begin{equation}
\mathcal{L}_{\text{ground}}
=
\frac{1}{2|\mathcal{P}|}
\sum_{(v,o)\in\mathcal{P}}
\left[
-\log
\frac{\exp(s_{v,o})}
{\sum_{o' \in \mathcal{O}_S} \exp(s_{v,o'})}
-
\log
\frac{\exp(s_{v,o})}
{\sum_{v' \in \mathcal{V}_S} \exp(s_{v',o})}
\right],
\end{equation}
where $\mathcal{O}_S$ and $\mathcal{V}_S$ denote the sets of grounded objects and views in a scene. 

\noindent\textbf{View-level and Scene-level Alignment.}
We further follow the alignment strategy of \cite{mao2025poma} to align the 3D encoder with the FG-CLIP~\cite{xie2025fg} text encoder at both view and scene levels using all scenes within the training batch. 

For view-level alignment, given $N_v$ views in a batch, the similarity logit between the RGB-Pointmap embedding $\mathbf{h}_v$ of view $v$ and the corresponding view caption embedding $\mathbf{t}_u^{V}$ of view $u$ is defined as $k_{v,u} = \mathbf{h}_v^\top \mathbf{t}_u^{V}/\tau$. Only matched pointmap–caption pairs are treated as positives, while all other combinations within the batch serve as negatives.
The view-level alignment objective can be formulated as:
\begin{equation}
\mathcal{L}_{\text{view}}
=
\frac{1}{2N_v}
\sum_{v=1}^{N_v}
\left[
-\log
\frac{\exp\!\left(k_{v,v}\right)}
{\sum_{u=1}^{N_v}\exp\!\left(k_{v,u}\right)}
-\log
\frac{\exp\!\left(k_{v,v}\right)}
{\sum_{u=1}^{N_v}\exp\!\left(k_{u,v}\right)}
\right].
\end{equation}

For scene-level alignment, the per-view RGB-Pointmap embeddings within a scene $\{\mathbf{h}_v\}_{v=1}^{N}$ are mean-pooled to obtain a scene-level embedding $\bar{\mathbf{h}}_s$. Given $N_s$ scenes in a batch, the similarity logit between the scene-level embedding $\bar{\mathbf{h}}_i$ of scene $i$ and the scene caption embedding $\mathbf{t}_j^S$ of scene $j$ is defined as $m_{i,j} = \bar{\mathbf{h}}_i^\top \mathbf{t}_j^S/\tau$. We align $\bar{\mathbf{h}}_i$ with its corresponding caption $\mathbf{t}_j^S$ using the following objective:
\begin{equation}
\mathcal{L}_{\text{scene}}
=
\frac{1}{2N_s}
\sum_{i=1}^{N_s}
\left[
-\log
\frac{\exp\!\left(m_{i,i}\right)}
{\sum_{j=1}^{N_s}\exp\!\left(m_{i,j}\right)}
-\log
\frac{\exp\!\left(m_{i,i}\right)}
{\sum_{j=1}^{N_s}\exp\!\left(m_{j,i}\right)}
\right].
\end{equation}

\noindent\textbf{Total Loss.} The total loss for UniScene3D pretraining is defined as below:
\begin{equation}
    \mathcal{L}_{\text{total}} = 
    \lambda\mathcal{L}_{\text{geo}} +
    \mathcal{L}_{\text{ground}} +
    \mathcal{L}_{\text{view}} + \mathcal{L}_{\text{scene}}. 
\end{equation} 
where $\lambda = 0.1$ in our setting. Together, these four alignment losses encourage the learning of geometrically consistent and semantically rich RGB-Pointmap features across multiple scales.

\section{Experiments}
\label{sec:results}
\subsection{Experimental Setting}
\noindent\textbf{Evaluation Tasks.} We evaluate the representation quality of UniScene3D on diverse 3D scene understanding tasks, including viewpoint grounding, scene retrieval, scene type classification, and 3D visual question answering (3D VQA). Viewpoint grounding assesses fine-grained spatial understanding by requiring the model to identify the view that best matches a referring expression. Scene retrieval aims to retrieve the correct 3D scene from descriptions at different levels of granularity. Scene type classification categorizes indoor environments into room types. These tasks are evaluated in zero-shot or few-shot settings to measure representation quality under minimal supervision. Such low-shot evaluation is common for vision-language encoders~\cite{radford2021learning,tschannen2025siglip}. 3D visual question answering (3D VQA) evaluates a model’s ability to answer natural language questions about a 3D scene by reasoning over its spatial layout, object relationships, and semantic attributes. We evaluate UniScene3D on this task under task-specific fine-tuning to examine how effectively it serves as a backbone for downstream 3D models.

\noindent\textbf{Datasets.} We pretrain UniScene3D on 6,562 indoor scenes from ScanNet~\cite{dai2017scannet}, 3RScan~\cite{wald2019rio}, and ARKitScenes~\cite{baruch2021arkitscenes}. For grounded view alignment, we use LLM-generated referring expressions from SceneVerse~\cite{jia2024sceneverse} and adopt LLM-generated view-level and scene-level captions from POMA-3D~\cite{mao2025poma}.

For evaluation, we reformulate existing visual grounding benchmarks, including ScanRefer~\cite{chen2020scanrefer}, Nr3D, and Sr3D~\cite{achlioptas2020referit3d}, to support viewpoint grounding and scene retrieval. Scene type classification is performed over 21 ScanNet categories (e.g., living room and bedroom). For 3D VQA, we evaluate on ScanQA~\cite{azuma2022scanqa} for commonsense spatial reasoning, SQA3D~\cite{ma2022sqa3d} for situated reasoning, and Hypo3D~\cite{mao2025hypo3d} for hypothetical reasoning. We follow the standard train–test splits used in prior work~\cite{zhu20233d,jia2024sceneverse,mao2025poma}. All evaluation code and datasets will be publicly released.

\noindent\textbf{Baselines.} We compare UniScene3D with diverse state-of-the-art perception encoders spanning different input modalities for 3D representation learning. For image-based methods, we include DFN~\cite{fang2023data}, SigLIP2~\cite{tschannen2025siglip}, and FG-CLIP~\cite{xie2025fg}, which take multi-view images as input. All models use ViT-B/16 checkpoints pretrained at 224$\times$224 resolution, matching the architecture of UniScene3D. For point cloud-based approaches, we evaluate Uni3D-g (Giant)~\cite{zhouuni3d}, 3D-VisTA~\cite{zhu20233d}, and SceneVerse~\cite{jia2024sceneverse}. As Uni3D-g is pretrained on small object-level point clouds (10,000 points), directly applying it to full-scene point clouds degrades performance. We therefore convert multi-view pointmaps into per-view point clouds, extract view-level embeddings using Uni3D-g, and aggregate them into a scene representation. For pointmap-based methods, we compare with POMA-3D~\cite{mao2025poma}, which is most closely related to our approach.

\subsection{Implementation Details. }We implement UniScene3D in PyTorch and train it on NVIDIA A100 GPUs. The model is pretrained for 80 epochs with a batch size of 64, using the ViT-B/16 variant of FG-CLIP~\cite{xie2025fg} as the backbone. We optimize the model using AdamW~\cite{loshchilov2017decoupled} with $\beta_1=0.9$ and $\beta_2=0.98$. The learning rate is initialized at $1\times10^{-4}$ and decayed to a minimum of $1\times10^{-5}$ following a cosine schedule. For each 3D scene, we sample 32 image–pointmap pairs using the maximum coverage sampling strategy from Video-3D-LLM~\cite{zheng2025video} and resize them to $224 \times 224$. This strategy selects a limited set of views that maximally cover the scene. For cross-view geometric alignment, we set $\alpha = 0.7$ and $\tau_r = 0.35$. See the Appendix for more details about downstream evaluation. 

\input{tables/view_retrieval}

\input{figures/view_retrieval_visualization}

\subsection{Viewpoint Grounding}
\noindent\textbf{Setting.} Given a referring expression describing an object, viewpoint grounding aims to identify the view within a 3D scene that best observes the referred object. We reformulate existing 3D visual grounding benchmarks by converting object-level annotations (object IDs and 3D bounding boxes) into view-level labels. Since an object may appear in multiple views, the ground-truth view is defined as the one where the object occupies the largest visible area. During evaluation, embeddings are computed for all views in a scene and compared with the referring text embedding. A prediction is considered correct if the most similar view matches the ground-truth view. Performance is reported using Recall@1 (R@1), Recall@5 (R@5), and Recall@10 (R@10). R@5 and R@10 additionally reflect how well the model identifies coarse viewpoints of the referred object.

\noindent\textbf{Results.} As shown in Table~\ref{tab:view_retrieval}, UniScene3D consistently outperforms all baselines across metrics, achieving R@1 gains of at least 15\% on ScanRefer, 8\% on Nr3D, and 10\% on Sr3D. In contrast, Uni3D-g performs worst, suggesting that object-level point cloud pretraining transfers poorly to scene-level reasoning and highlighting the need for 3D encoders designed specifically for scene understanding. Notably, the prior pointmap-based method POMA-3D underperforms image-based encoders, despite the task emphasizing 3D grounding and inter-object spatial reasoning that largely depend on geometric cues rather than appearance. UniScene3D nearly doubles the R@1 of POMA-3D on most benchmarks, indicating that our training strategy more effectively leverages geometric information from pointmaps for spatial reasoning.

The qualitative results in Fig.~\ref{fig:view_retrieval_visualization} further support these findings. UniScene3D successfully retrieves the correct views in large, cluttered scenes requiring either complex appearance reasoning (top) or geometric reasoning (bottom).

\input{tables/scene_retrieval}

\subsection{Scene Retrieval}
\noindent\textbf{Setting.} The scene retrieval task aims to retrieve the scene that best matches a given caption by computing the similarity between scene embeddings and the caption embedding. Scene-level baselines (3D-VisTA and SceneVerse) directly produce scene embeddings, while for view-based methods, we obtain scene embeddings by mean-pooling view-level embeddings. Following POMA-3D~\cite{mao2025poma}, scene captions are constructed by concatenating referring texts from 3D visual grounding datasets. We evaluate under two settings where each caption contains ($n=5$) or ($n=10$) texts, and report R@1 and R@5.

\noindent\textbf{Results. }Table~\ref{tab:scene_retrieval} demonstrates that UniScene3D achieves a clear performance advantage on this task, with results further improving as longer and more detailed scene captions are provided. In contrast, point cloud-based methods perform poorly, sometimes even falling behind image-based encoders such as DFN and FG-CLIP. This observation supports our assumption that training point cloud encoders from scratch limits their ability to exploit rich 2D foundation priors and increases the risk of overfitting to training descriptions. By contrast, pointmap-based methods that are fine-tuned from pretrained 2D encoders, including POMA-3D and UniScene3D, exhibit stronger generalization.

\subsection{Scene Type Classification}
\noindent\textbf{Setting. }Scene type classification evaluates a model’s ability to predict the correct category of a 3D scene. We assess this task in both zero-shot and few-shot settings. In the zero-shot setting, scene embeddings are matched against 21 candidate scene types using the prompt template ``This room is a \{\}.'' A prediction is correct if the scene embedding has the highest similarity to the corresponding scene-type prompt. In the few-shot setting, we perform linear probing with L-BFGS~\cite{liu1989limited} using $N$ labeled examples per class ($N$ shots) and evaluate on the remaining unseen scenes.
\input{tables/scene_classification}
\input{tables/3d_vqa_specialist}

\noindent\textbf{Results. }Table~\ref{tab:scene_classification} shows that UniScene3D ranks first across all settings. Point cloud–based methods perform the worst. Surprisingly, scene-level point cloud methods even underperform the object-level model Uni3D in the zero-shot setting. This may be because Uni3D is pretrained on much larger object-level point clouds than those available for scene-level models, leading to weaker zero-shot generalization for scene-level approaches. Other baselines exhibit fluctuating performance and only excel under specific supervision regimes. For example, POMA-3D performs best among baselines in the 0- and 5-shot settings but falls behind image-based methods under 1- and 10-shot evaluation. DFN achieves the strongest 1-shot result, while SigLIP2 performs best at 10-shot. Such variability suggests limited robustness across supervision levels. In contrast, UniScene3D maintains consistently high performance across all settings.
\input{tables/design_choice}

\subsection{3D Visual Question Answering}
\noindent\textbf{Setting. }Following prior work~\cite{jia2024sceneverse,mao2025poma}, we evaluate 3D VQA by attaching a shallow QA head and a BERT~\cite{koukounas2024jina} language encoder. The visual encoder is frozen, while only the language encoder and QA head are fine-tuned. We report Exact Match at top-1 and top-10 (EM@1, EM@10) as evaluation metrics.

\noindent\textbf{Results. }Consistent with the findings on other downstream tasks, Table~\ref{tab:3d_vqa_specialist} shows that UniScene3D consistently outperforms all 3D encoder baselines across the three benchmarks. Notably, POMA-3D underperforms image-based methods on ScanQA but surpasses them on SQA3D and Hypo3D. This discrepancy may stem from the fact that ScanQA contains many questions involving color and texture cues. By incorporating image appearance information, UniScene3D effectively enhances pointmap-based representations on this benchmark.

\subsection{Ablation Study}
\input{tables/ablation}
\noindent\textbf{2D Foundation Priors Benefit Pointmap Learning.} Our work hypothesizes that the grid structure of pointmaps enables effective transfer of pretrained 2D priors. To validate this, we construct a pointmap-only baseline using the same ViT backbone as FG-CLIP, initialized with either random weights or FG-CLIP pretrained weights. On viewpoint grounding (ScanRefer), the pretrained model achieves consistent improvement over random initialization, even without further training on 3D scenes. After applying the full UniScene3D training strategy, the performance gap widens further. These results provide direct evidence that 2D foundation priors significantly improve pointmap representation learning, both as a standalone encoder and as a strong initialization for subsequent training.

\noindent\textbf{Effect of RGB-Pointmap Representation.} Table~\ref{tab:ablation_three_tasks} compares UniScene3D with two ablated variants: a pointmap-only model (w/o image input) and an image-only model (w/o pointmap input), both removing the early token fusion mechanism and reducing the architecture to a standard ViT. UniScene3D consistently outperforms these baselines across all zero-shot and few-shot tasks, demonstrating that the proposed RGB-Pointmap representation effectively integrates complementary geometric and appearance cues to learn stronger unified 3D scene representations. Qualitative examples in Fig.~\ref{fig:figure1} further illustrate this advantage: the pointmap-only variant fails to identify green seats due to missing color cues, while the image-only variant cannot recognize the longest seating area without geometric information. These results highlight the necessity of jointly modeling appearance and geometry for robust 3D scene understanding.

We further compare our early token fusion strategy with alternative design choices in Table~\ref{tab:scanrefer}. When the patch embedding is randomly initialized instead of using the pretrained image patch embedding layer, or when the positional encoding is removed, performance drops noticeably. This validates that preserving pretrained 2D weights is important for effective RGB-Pointmap learning. We also evaluate a simple fusion strategy that concatenates image and pointmap inputs, followed by a projection layer to match the encoder dimension. This strategy performs substantially worse than our design and even underperforms the w/o image setting in Table~\ref{tab:ablation_three_tasks}. This may be because the projection layer has to be trained from scratch, which harms the generalization of the fused features.

\noindent\textbf{Effect of Training Strategy.} Table~\ref{tab:ablation_three_tasks} shows that both cross-view geometric alignment and grounded view alignment contribute to learning effective features in UniScene3D across downstream tasks. Cross-view geometric alignment enforces global geometric consistency and, even without explicit semantic alignment with text, still yields consistent improvements on cross-modal grounding and retrieval tasks. Grounded view alignment further enhances the model’s grounding capability, producing particularly strong gains on the viewpoint grounding task. These results suggest that effective pointmap embeddings should capture both geometric structure and semantic relationships within a scene.

\noindent\textbf{Number of Views Analysis.} We further show that UniScene3D is robust to the number of input views in Fig.~\ref{fig:number_view}, consistently outperforming POMA-3D across all settings. Notably, using only 16 views already surpasses POMA-3D with 32 views, indicating that our pretraining does not bias the model toward a specific number of viewpoints and supports broader applicability.

\noindent\textbf{Data Scaling Analysis.}  Fig.~\ref{fig:data_ratio} shows that UniScene3D scales positively with more training scenes, with performance improving steadily as the dataset size increases. Training on the full dataset clearly outperforms using only half of the data. While UniScene3D benefits from pretrained 2D priors, it remains data-hungry. We expect further gains with larger-scale real-world pretraining data.

\section{Limitation} Although UniScene3D learns effective 3D representations for a wide range of downstream tasks, it still has several limitations. First, UniScene3D is pretrained with a fixed input resolution of image–pointmap pairs and a ViT-B/16 backbone. Pretraining with varied input resolutions and model sizes would be necessary for UniScene3D to scale as a general-purpose 3D representation framework across diverse scenarios. Second, UniScene3D is trained on a limited number of scenes. Although it benefits from strong 2D foundation model priors, larger-scale pretraining on more diverse 3D scenes could further improve its effectiveness.

\section{Conclusion}
In this paper, we present UniScene3D, a 3D representation learning method that jointly models geometry and appearance from multi-view RGB-Pointmap. Leveraging the rich cross-view context information pointmaps, we introduce cross-view geometric alignment and grounded view alignment objectives for pretraining. Experiments show that UniScene3D achieves state-of-the-art performance across multiple 3D scene understanding benchmarks while remaining robust to varying numbers of input views. In future work, we plan to scale UniScene3D to larger model sizes, similar to 2D CLIP models, and explore its use as a general vision backbone for building 3D foundation models.

\bibliographystyle{splncs04}
\bibliography{main}
\end{document}

%% file: figures/teaser.tex
{
\renewcommand\twocolumn[1][]{#1}
\thispagestyle{empty}
    \begin{center}
       \includegraphics[width=1.0\linewidth,page=1]{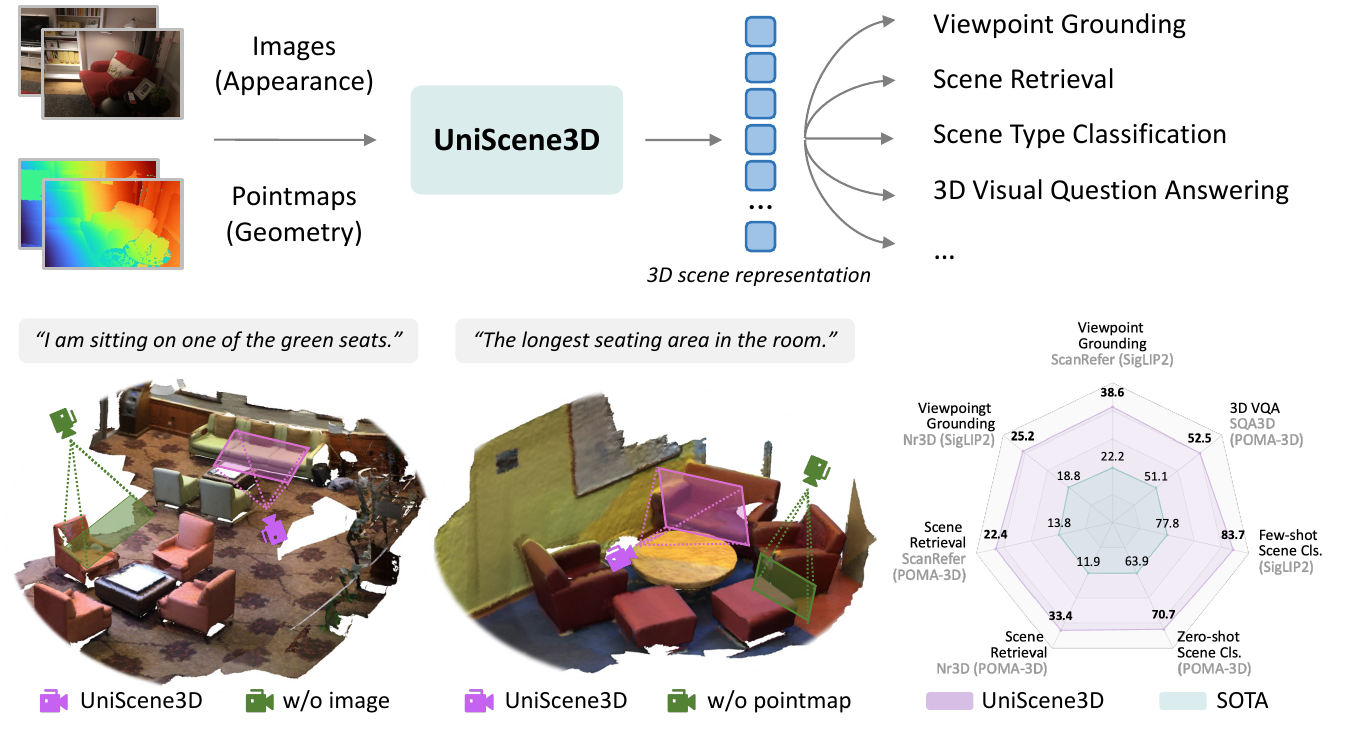}
        \vspace{-1em}
        \captionof{figure}{\textbf{Overview of UniScene3D.}
        \emph{Top:} UniScene3D takes multi-view images and pointmaps as input to learn 3D representations for viewpoint grounding, scene retrieval, zero-/few-shot scene type classification, and 3D visual question answering. 
        \emph{Bottom:} Example of viewpoint grounding. Image appearance cues enable correct color recognition (left), while pointmap geometry supports reasoning about spatial extent, enabling identification of the longest seat (right). The radar chart shows comparisons between UniScene3D and prior state-of-the-art methods across multiple tasks and benchmarks.}
        \label{fig:figure1}
        \vspace{-0.4em}
    \end{center}
}

%% file: figures/uniscene3d.tex
\definecolor{cyan}{RGB}{180, 210, 220}
\begin{figure*}[!t]
    \centering
    \includegraphics[width=1.0\linewidth]{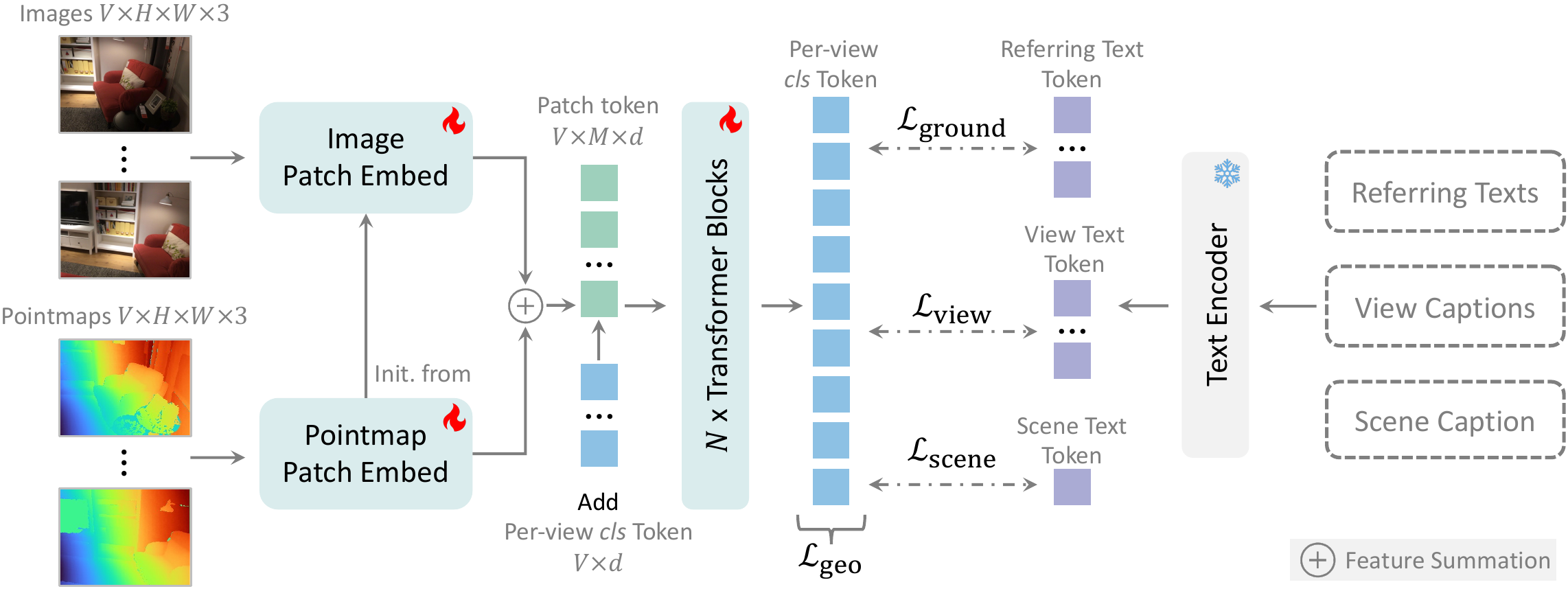}
    \vspace{-1.5em}
    \caption{\textbf{Overview of UniScene3D pretraining.} UniScene3D takes multi-view image–pointmap pairs as input and performs early fusion at the patch embedding stage. The fused tokens, added with absolute positional encodings, are then processed by $N$ Transformer blocks to produce a unified RGB-Pointmap representation. During pretraining, UniScene3D is optimized with four alignment objectives: (1) \emph{Cross-view geometric alignment} $\mathcal{L}_{\text{geo}}$; (2) \emph{Grounded view alignment} $\mathcal{L}_{\text{ground}}$; (3) \emph{View-level alignment} $\mathcal{L}_{\text{view}}$; and (4) \emph{Scene-level alignment} $\mathcal{L}_{\text{scene}}$. Its blocks are highlighted in \textcolor{cyan}{cyan}.}
    \label{fig:2}
    \vspace{-0.5em}
\end{figure*}

%% file: tables/comparison_of_alignment.tex
\begin{table}[t]
\addtolength{\tabcolsep}{4.0pt}
\centering
\scriptsize
\caption{\textbf{Alignment strategies in UniScene3D.} RGBP denotes RGB-Pointmap embeddings. Contrastive pairs are sampled either from views within the same scene or from the entire training batch, including different scenes. Cross-view geometric alignment operates on RGBP--RGBP pairs, while the others perform RGBP--Text alignment. Scene-level alignment uses aggregated view embeddings (Scene embedding); the remaining ones operate on per-view embeddings.}
\vspace{-1em}
\begin{tabular}{c|ccc|c}
\toprule
Alignment Strategy
& Contrastive Pairs 
& Modalities 
& Embedding 
& Target \\
\midrule
Cross-view geometric 
& Scene
& RGBP--RGBP 
& Per-view 
& Geometry consistency \\

Grounded view 
& Scene 
& RGBP--Text 
& Per-view 
& Object semantics \\

View-level 
& Batch
& RGBP--Text 
& Per-view 
& View semantics \\

Scene-level 
& Batch 
& RGBP--Text 
& Scene
& Scene semantics \\
\bottomrule
\end{tabular}
\label{tab:2}
\vspace{-0.5em}
\end{table}

%% file: tables/view_retrieval.tex
\begin{table*}[!t]
\addtolength{\tabcolsep}{4.0pt}
\scriptsize
\centering
\caption{\textbf{Viewpoint grounding results on ScanRefer, Nr3D, and Sr3D.} Input modalities include multi-view image (I), Point cloud (PC), Pointmap (PM), and RGB-Pointmap (RGBP). R@N (Recall@N) measures the percentage of cases where the correct view is ranked within the top $N$ candidates retrieved. All methods are evaluated in the zero-shot setting. The best results are in \textbf{bold}.}
\vspace{-1em}
\begin{tabular}{l|c|ccc|ccc|ccc}
\toprule
\multirow{2}{*}[-0.7ex]{Method} & 
\multirow{2}{*}[-0.7ex]{Input} &
\multicolumn{3}{c|}{ScanRefer} &
\multicolumn{3}{c|}{Nr3D} &
\multicolumn{3}{c}{Sr3D} \\
\cmidrule(lr){3-5} \cmidrule(lr){6-8} \cmidrule(lr){9-11}
& & 
R@1 & R@5 & R@10 &
R@1 & R@5 & R@10 &
R@1 & R@5 & R@10 \\
\midrule
DFN~\cite{fang2023data} & I &
 20.9 & 59.2 & 77.6 &
 17.3 & 51.8 & 71.8 &
 12.6 & 43.5 & 65.6 \\
SigLIP2~\cite{tschannen2025siglip} & I &
 22.2 & 61.1 & 79.5 &
 18.8 & 54.3 & 74.0 &
 13.0 & 44.4 & 67.2 \\
FG-CLIP~\cite{xie2025fg} & I &
 20.2 & 58.5 & 77.6 &
 17.3 & 52.6 & 72.8 &
 13.4 & 44.5 & 66.6 \\
Uni3D-g~\cite{zhouuni3d} & PC &
 4.2 & 19.0 & 35.5 &
 3.6 & 17.2 & 33.4 &
 3.4 & 16.5 & 32.2 \\
POMA-3D~\cite{mao2025poma} & PM &
 16.4 & 50.6 & 71.2 &
 13.6 & 44.5 & 65.9 &
 10.2 & 37.4 & 59.2 \\
UniScene3D & RGBP &
 \textbf{38.6} & \textbf{72.5} & \textbf{85.7} &
 \textbf{25.2} & \textbf{64.0} & \textbf{80.9} &
 \textbf{23.6} & \textbf{62.8} & \textbf{80.4} \\
\bottomrule
\end{tabular}
\label{tab:view_retrieval}
\vspace{-1em}
\end{table*}

%% file: figures/view_retrieval_visualization.tex
\definecolor{forestgreen}{RGB}{34,139,34}
\begin{figure}[t]
  \centering
  \includegraphics[width=\linewidth]{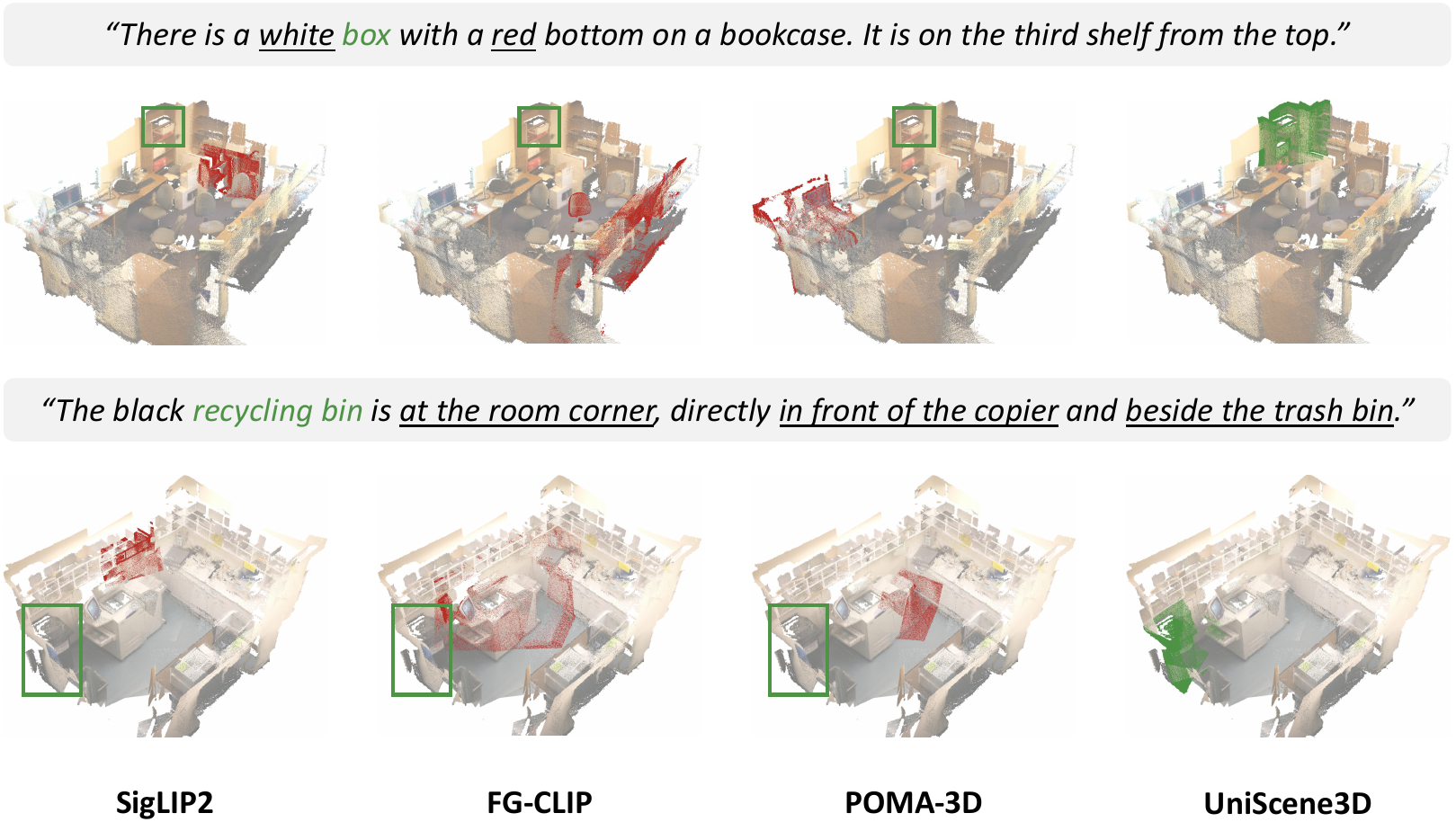}
  \vspace{0em}
  \caption{\textbf{Qualitative viewpoint grounding results.} Correct and incorrect matches are highlighted in \textcolor{forestgreen}{green} and \textcolor{red}{red}, respectively, based on the referring texts. Underlined phrases denote contextual clues that guide the models in solving the grounding task.}
  \label{fig:view_retrieval_visualization}
\end{figure}

%% file: tables/scene_retrieval.tex
\begin{table*}[!t]
\addtolength{\tabcolsep}{1.4pt}
\scriptsize
\caption{\textbf{Scene retrieval results on ScanRefer, Nr3D, and Sr3D.} We report results where each scene caption contains $n = 5$ or $10$ utterances. All methods are evaluated in the zero-shot setting.}
\vspace{-1em}
\begin{tabular}{l|c|cccc|cccc|cccc}
\toprule
\multirow{2}{*}[-2.8ex]{Method} &
\multirow{2}{*}[-2.8ex]{Input} &
\multicolumn{4}{c|}{ScanRefer} &
\multicolumn{4}{c|}{Nr3D} &
\multicolumn{4}{c}{Sr3D} \\
\cmidrule(lr){3-6} \cmidrule(lr){7-10} \cmidrule(lr){11-14}
& &
\multicolumn{2}{c}{$n=5$} & \multicolumn{2}{c|}{$n=10$} &
\multicolumn{2}{c}{$n=5$} & \multicolumn{2}{c|}{$n=10$} &
\multicolumn{2}{c}{$n=5$} & \multicolumn{2}{c}{$n=10$} \\
\cmidrule(lr){3-4} \cmidrule(lr){5-6}
\cmidrule(lr){7-8} \cmidrule(lr){9-10}
\cmidrule(lr){11-12} \cmidrule(lr){13-14}
& &
R@1 & R@5 & R@1 & R@5 &
R@1 & R@5 & R@1 & R@5 &
R@1 & R@5 & R@1 & R@5 \\
\midrule
DFN~\cite{fang2023data} 
& I & 4.6 & 13.5 & 4.6 & 13.3
& 3.9 & 12.7 & 4.4 & 13.0
& 0.6 & 3.1 & 0.6 & 3.2 \\
SigLIP2~\cite{tschannen2025siglip} 
& I & 1.5 & 5.3 & 1.8 & 6.6
& 0.7 & 3.0 & 1.0 & 3.8
& 0.1 & 0.3 & 0.1 & 0.4 \\
FG-CLIP~\cite{xie2025fg} 
& I & 9.2 & 26.3 & 16.3 & 40.0
& 6.2 & 20.8 & 12.4 & 32.8
& 0.6 & 2.7 & 0.7 & 3.8 \\
Uni3D-g~\cite{zhouuni3d}
& PC & 0.3 & 1.4 & 0.3 & 1.3
& 0.5 & 1.4 & 0.4 & 1.5
& 0.3 & 1.6 & 0.2 & 1.4 \\
3D-VisTA~\cite{zhu20233d}
& PC & 0.6 & 2.4 & 0.7 & 2.2
& 0.1 & 0.9 & 0.2 & 1.3
& 0.2 & 1.5 & 0.2 & 1.7 \\
SceneVerse~\cite{jia2024sceneverse}
& PC & 0.5 & 2.1 & 0.6 & 2.7
& 0.5 & 1.3 & 0.5 & 1.4
& 0.4 & 1.8 & 0.1 & 1.9 \\
POMA-3D~\cite{mao2025poma}
& PM & 13.8 & 34.9 & 20.4 & 47.2
& 11.9 & 32.7 & 19.3 & 46.8
& 1.6 & 6.9 & 2.4 & 9.5 \\
UniScene3D
& RGBP & \textbf{22.4} & \textbf{48.6} & \textbf{33.4} & \textbf{62.9}
& \textbf{19.7} & \textbf{46.3} & \textbf{30.7} & \textbf{61.7}
& \textbf{3.0} & \textbf{11.8} & \textbf{4.6} & \textbf{17.5} \\
\bottomrule
\end{tabular}
\label{tab:scene_retrieval}
\end{table*}

%% file: tables/scene_classification.tex
\begin{table}[!t]
    \addtolength{\tabcolsep}{0.1pt}
    \centering
    \scriptsize
    \begin{minipage}[t]{0.52\textwidth}
    \centering
    \caption{\textbf{Scene type classification accuracy (\%).} Linear probing is used for 1/5/10-shot evaluation. Our method consistently outperforms other methods.}
    \vspace{-0.3em}
    \begin{tabular}{l|c|cccc}
    \toprule
    Method & Input & 0-shot & 1-shot & 5-shot & 10-shot \\
    \midrule
    DFN~\cite{fang2023data} & I & 41.6 & 40.5 & 60.1 & 77.5 \\
    SigLIP2~\cite{tschannen2025siglip} & I & 36.0 & 36.6 & 57.8 & 77.9 \\
    FG-CLIP~\cite{xie2025fg} & I & 49.5 & 40.0 & 62.0 & 77.5 \\
    Uni3D-g~\cite{zhouuni3d} & PC & 8.6 & 21.4 & 37.1 & 45.4 \\
    3D-VisTA~\cite{zhu20233d} & PC & 1.9 & 28.9 & 49.4 & 55.0 \\
    SceneVerse~\cite{jia2024sceneverse} & PC & 0.8 & 24.0 & 51.1 & 64.2 \\
    POMA-3D~\cite{mao2025poma} & PM & 63.9 & 32.2 & 64.9 & 74.1 \\
    UniScene3D & RGBP & \textbf{70.7} & \textbf{43.2} & \textbf{72.4} & \textbf{83.7} \\
    \bottomrule
    \end{tabular}
    \label{tab:scene_classification}
    \end{minipage}%
    \hfill
    \begin{minipage}[t]{0.46\textwidth}
    \centering
    \captionsetup{type=figure}
    \includegraphics[width=\linewidth]{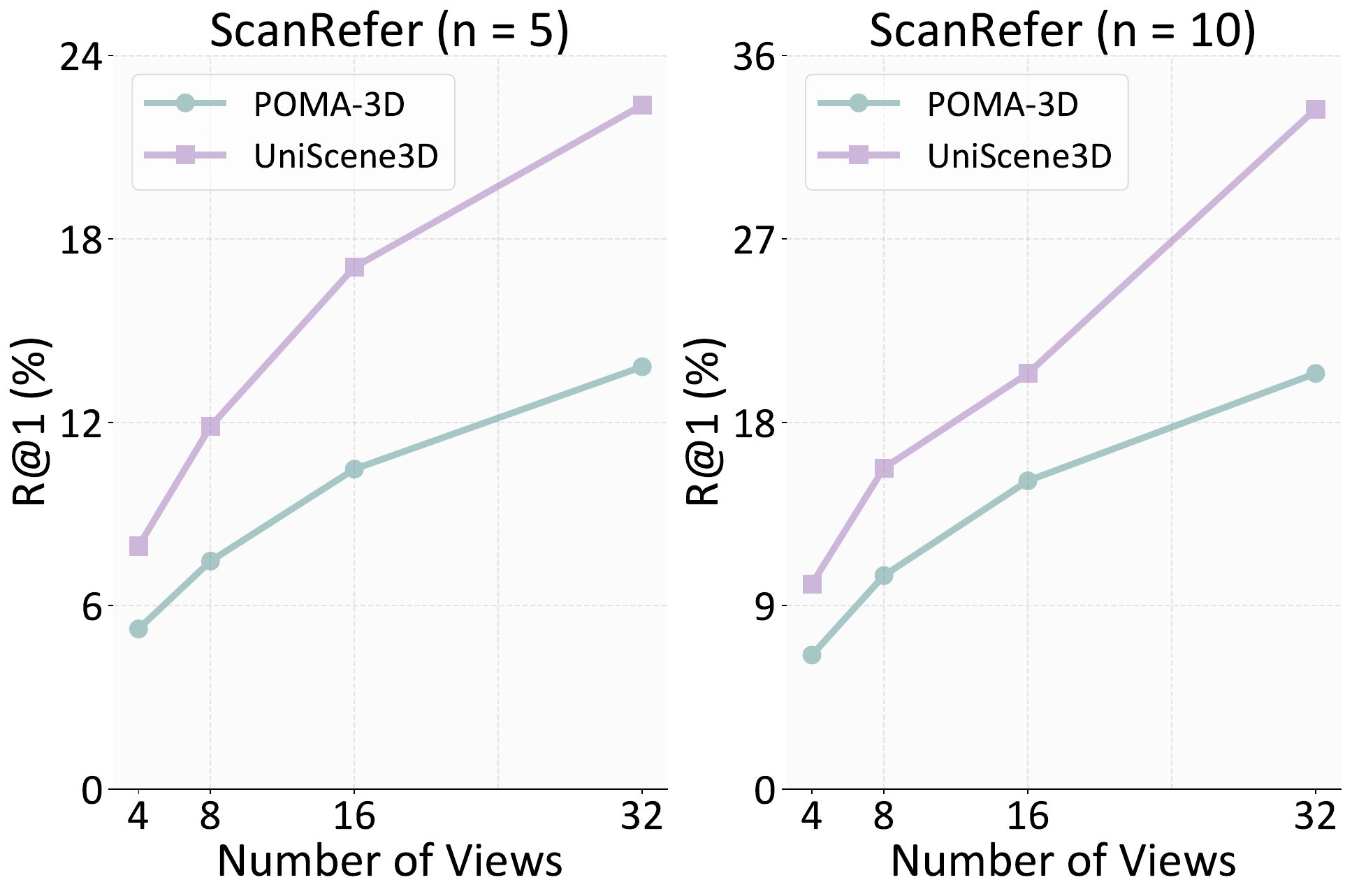}
    \caption{\textbf{Effect of view number on scene retrieval (ScanRefer).} R@1 is reported under $n=5$ and $n=10$.}
    \label{fig:number_view}
    \end{minipage}
    \vspace{-1em}
\end{table}

%% file: tables/3d_vqa_specialist.tex
\begin{table}[t]
    \addtolength{\tabcolsep}{6.5pt}
    \centering
    \scriptsize
    \caption{\textbf{3D VQA results on ScanQA, SQA3D, and Hypo3D.} We report Exact Match (EM@1/10) metrics. Higher value is better for all metrics.}
    \vspace{-1em}
    \begin{tabular}{l | c | cc | cc | cc}
    \toprule
    \multirow{2}{*}[-0.7ex]{Method} & 
    \multirow{2}{*}[-0.7ex]{Input} &
    \multicolumn{2}{c|}{ScanQA} &
    \multicolumn{2}{c|}{SQA3D} &
    \multicolumn{2}{c}{Hypo3D} \\
    
    \cmidrule(lr){3-4}
    \cmidrule(lr){5-6}
    \cmidrule(lr){7-8}
    
    & & EM@1 & EM@10 & EM@1 & EM@10 & EM@1 & EM@10 \\
    \midrule
    
    FG-CLIP~\cite{xie2025fg}            
    & I
    & 20.9 & 49.9 
    & 49.5 & 89.7 
    & 31.1 & 82.1 \\
    
    3D-VisTA~\cite{zhu20233d}          
    & PC 
    & 22.4 & 52.1 
    & 48.5 & 85.6 
    & 31.0 & 81.2 \\
    
    SceneVerse~\cite{jia2024sceneverse}     
    & PC
    & 22.7 & 51.5 
    & 49.9 & 85.0 
    & 31.6 & 80.3 \\
    
    POMA-3D~\cite{mao2025poma}       
    & PM
    & 22.3 & 52.3 
    & 51.1 & 91.2 
    & 33.4 & 84.8 \\
    
    UniScene3D       
    & RGBP
    & \textbf{23.2} & \textbf{53.5} 
    & \textbf{52.5} & \textbf{92.3}
    & \textbf{35.2} & \textbf{85.9} \\
    
    \bottomrule
    \end{tabular}
    \label{tab:3d_vqa_specialist}
    \vspace{-0.5em}
\end{table}

%% file: tables/design_choice.tex
\begin{table*}[t]
\centering
\begin{minipage}[t]{0.53\textwidth}
    \centering
    \captionsetup{type=table}
    \caption{\textbf{Effect of initialization and pretraining on viewpoint grounding (ScanRefer).} 2D priors benefit pointmap learning.}
    \vspace{-1em}
    \scriptsize
    \addtolength{\tabcolsep}{4.5pt}
    \begin{tabular}{l|cc|cc}
    \toprule
    \multirow{2}{*}[-0.5ex]{Initialization} 
    & \multicolumn{2}{c|}{w/o Pretrain} 
    & \multicolumn{2}{c}{w/ Pretrain} \\
    \cmidrule(lr){2-3} \cmidrule(lr){4-5}
    & R@1 & R@5 & R@1 & R@5 \\
    \midrule
    Random             
    & 4.32 & 17.5 & 7.53 & 24.38 \\
FG-CLIP~\cite{xie2025fg}               
    & \textbf{5.03} & \textbf{21.1} & \textbf{37.62} & \textbf{71.53} \\
    \bottomrule
    \end{tabular}
    \label{tab:effect_of_initialization_and_pretraining}
    \vspace{2pt}
    \centering
    \captionsetup{type=table}
    \caption{\textbf{Ablations of image-pointmap (PM) fusion on viewpoint grounding (ScanRefer).} Init. denotes initialization.}
    \vspace{-1em}
    \scriptsize
    \addtolength{\tabcolsep}{0.6pt}
    \begin{tabular}{l|cc}
    \toprule
    Case & R@1 & R@5 \\
    \midrule
    UniScene3D & \textbf{38.6} & \textbf{72.5} \\
    Random Init. PM Patch Embed  & 38.0 & 71.8 \\
    w/o Positional Encoding & 37.2 & 71.4 \\
    Concat + Projection & 36.5 & 70.0 \\
    \bottomrule
    \end{tabular}
    \label{tab:scanrefer}
\end{minipage}%
\hfill
\begin{minipage}[t]{0.44\textwidth}
    \centering
    \captionsetup{type=figure}
    \includegraphics[width=\linewidth]{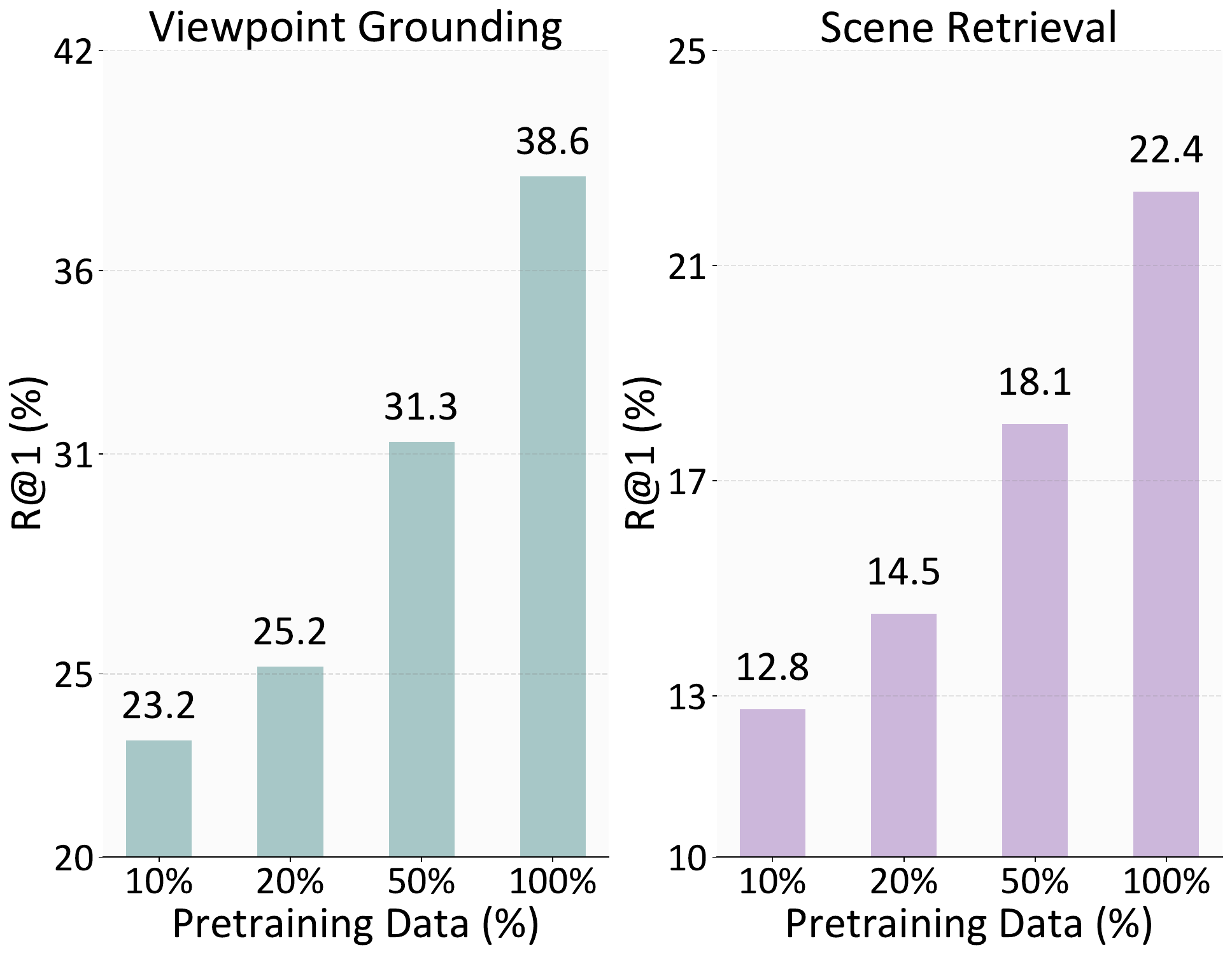}
    \caption{\textbf{Effect of pretraining data scale on viewpoint grounding and scene retrieval (ScanRefer, R@1, $n = 5$)}. Performance improves consistently with more pretraining data.}
    \label{fig:data_ratio}
\end{minipage}
\vspace{-1em}
\end{table*}

%% file: tables/ablation.tex
\begin{table*}[!t]
\centering
\scriptsize
\addtolength{\tabcolsep}{1.7pt}
\caption{\textbf{Ablation study of UniScene3D on viewpoint grounding, scene retrieval (ScanRefer), and scene type classification.} We compare variants with individual components removed against the full model.}
\vspace{-1em}
\begin{tabular}{l|cccccc|cccc|cc}
\toprule
\multirow{2}{*}[-3ex]{Case} &
\multicolumn{6}{c|}{Viewpoint Grounding} &
\multicolumn{4}{c|}{Scene Retrieval} &
\multicolumn{2}{c}{Scene Type} \\
\cmidrule(lr){2-7}
\cmidrule(lr){8-11}
& \multicolumn{2}{c}{ScanRefer}
& \multicolumn{2}{c}{Nr3D}
& \multicolumn{2}{c|}{Sr3D}
& \multicolumn{2}{c}{$n=5$}
& \multicolumn{2}{c|}{$n=10$}
& \multicolumn{2}{c}{Classification} \\
\cmidrule(lr){2-3}
\cmidrule(lr){4-5}
\cmidrule(lr){6-7}
\cmidrule(lr){8-9}
\cmidrule(lr){10-11}
\cmidrule(lr){12-13}
& R@1 & R@5
& R@1 & R@5
& R@1 & R@5
& R@1 & R@5
& R@1 & R@5
& 0-shot & 10-shot \\
\midrule
UniScene3D &
\textbf{38.6} & \textbf{72.5} &
\textbf{25.2} & \textbf{64.0} &
\textbf{23.6} & \textbf{62.8} &
\textbf{22.4} & \textbf{48.6} &
\textbf{33.4} & \textbf{62.9} &
\textbf{70.7} & \textbf{83.7} \\
\quad w/o image &
37.6 & 71.7 &
23.7 & 62.7 &
23.1 & 62.0 &
21.2 & 45.7 &
31.2 & 60.5 &
67.2 & 83.6 \\
\quad w/o pointmap &
28.8 & 66.5 &
21.3 & 58.3 &
18.5 & 54.9 &
21.8 & 48.0 &
32.0 & 61.9 &
68.9 & 82.6 \\
\quad w/o $\mathcal{L}_{\text{geo}}$ &
38.2 & 71.5 &
23.2 & 62.0 &
23.0 & 62.3 &
21.6 & 47.3 &
32.4 & 61.8 &
68.7 & 81.3 \\
\quad w/o $\mathcal{L}_{\text{ground}}$ &
18.5 & 56.3 &
15.6 & 49.7 &
11.5 & 41.5 &
21.8 & 47.6 &
30.6 & 63.0 &
68.5 & 82.4 \\
\bottomrule
\end{tabular}
\label{tab:ablation_three_tasks}
\end{table*}